\pdfoutput=1
\documentclass[10pt,twocolumn,letterpaper]{article}

\usepackage[utf8]{inputenc}
\usepackage[T1]{fontenc}
\usepackage{amsmath}
\usepackage{newtxtext,newtxmath}
\usepackage[margin=0.78in,top=0.82in,bottom=0.86in,columnsep=0.26in]{geometry}
\usepackage{graphicx}
\usepackage{booktabs}
\usepackage{multirow}
\usepackage{tabularx}
\usepackage{array}
\usepackage{xcolor}
\usepackage{colortbl}
\usepackage{microtype}
\usepackage{enumitem}
\usepackage{caption}
\usepackage{titlesec}
\usepackage[numbers,sort&compress]{natbib}
\usepackage[colorlinks=true,linkcolor=blue!65!black,citecolor=blue!65!black,urlcolor=blue!65!black]{hyperref}

\definecolor{bestbg}{HTML}{FDEDEC}
\definecolor{litebg}{HTML}{E8F8F5}
\definecolor{secblue}{HTML}{1B4F72}
\definecolor{darkred}{HTML}{922B21}

\titlespacing*{\section}{0pt}{1.7ex plus 0.4ex minus 0.2ex}{0.85ex plus 0.2ex}
\titlespacing*{\subsection}{0pt}{1.25ex plus 0.3ex minus 0.2ex}{0.55ex plus 0.1ex}

\newcommand{\best}[1]{\textcolor{darkred}{\textbf{#1}}}
\newcommand{\secbest}[1]{\underline{#1}}

\title{\vspace{-1.1em}\textbf{PhysRF-WorldJEPA: Transforming Heterogeneous Large Language Models into Manifold-Decoupled Millimeter-Wave Radar World Models}\vspace{-0.3em}}

\author{
  Yicheng Huang \hspace{0.9em} Feng Ke\thanks{Corresponding author: Feng Ke (\texttt{fengke@scut.edu.cn}).} \hspace{0.9em} Linhan Chen \hspace{0.9em} Ziyi Jiang \hspace{0.9em} Xiu Yin Zhang \\[0.25em]
  {\normalsize School of Electronic and Information Engineering, South China University of Technology, Guangzhou, China}
}

\date{\vspace{-1.2em}}

\begin{document}
\maketitle

\begin{abstract}
Adapting Large Language Models (LLMs) to millimeter-wave (mmWave) radar enables privacy-preserving, all-weather physical scene understanding across darkness, smoke, and walls. However, standard instruction tuning treats continuous radar streams as static text-like tokens, causing near-random global radar-to-text retrieval and severe physical drift under sensor blackouts. Conversely, coupling pairwise contrastive losses (e.g., SigLIP) directly into the shared backbone improves retrieval but injects $O(B^2)$ repulsive gradients that fracture continuous coordinate manifolds ($R^2 < 0$). We present \textbf{PhysRF-WorldJEPA}, a manifold-decoupled Joint-Embedding Predictive Architecture (JEPA) that separates continuous electromagnetic state prediction from cross-modal language alignment. An unattenuated $1.0\times$ clock anchor and parallel dual-scale RF encoder feed a lean $\sim\!3\%$ LoRA LLM trained with two-step state-space EMA-JEPA ($\text{Denoise}\!\rightarrow\!\text{Rollout}$) and sample-axis Batch-VICReg to predict $+3.0\,\text{s}$ physical states; a stop-gradient alignment head then maps frozen prompt states onto the text hypersphere with false-negative masking. On a 4,541-window tri-domain suite ($77/60.25/60\,\text{GHz}$) against 10 baselines, freezing the physical backbone and training only the decoupled head lifts $\text{R}\!\rightarrow\!\text{T}$ R@10 from $5.95\%$ to $\mathbf{15.01\%}$ and improves MedRank from $107.0$ to $\mathbf{60.0}$ while preserving identical physical readout: $3.37^\circ$ ($R^2\!=\!+0.961$) blackout $+3.0\,\text{s}$ torso tilt, $1.89\,\text{m}$ ($R^2\!=\!+0.679$; $1.19\,\text{m}$/$+0.863$ at the RF front-end) cluttered UAV altitude, $2.04\,\text{m}$ ($R^2\!=\!+0.461$) behind-wall NLOS range, $0.9976$ virtual-RIS invariance, and $100\%$ fall recall. Across five $0.5\text{B}\text{--}1.8\text{B}$ LLM families (15 zero-overhead models), World-JEPA improves retrieval and $75\%$-drop completion in $10/10$ runs and blackout tilt in $4/5$ (Span-Euc), $2/5$ (Scatter), and $5/5$ (per-family best) over SFT.
\end{abstract}

\begin{figure*}[t]
  \centering
  \includegraphics[width=\textwidth]{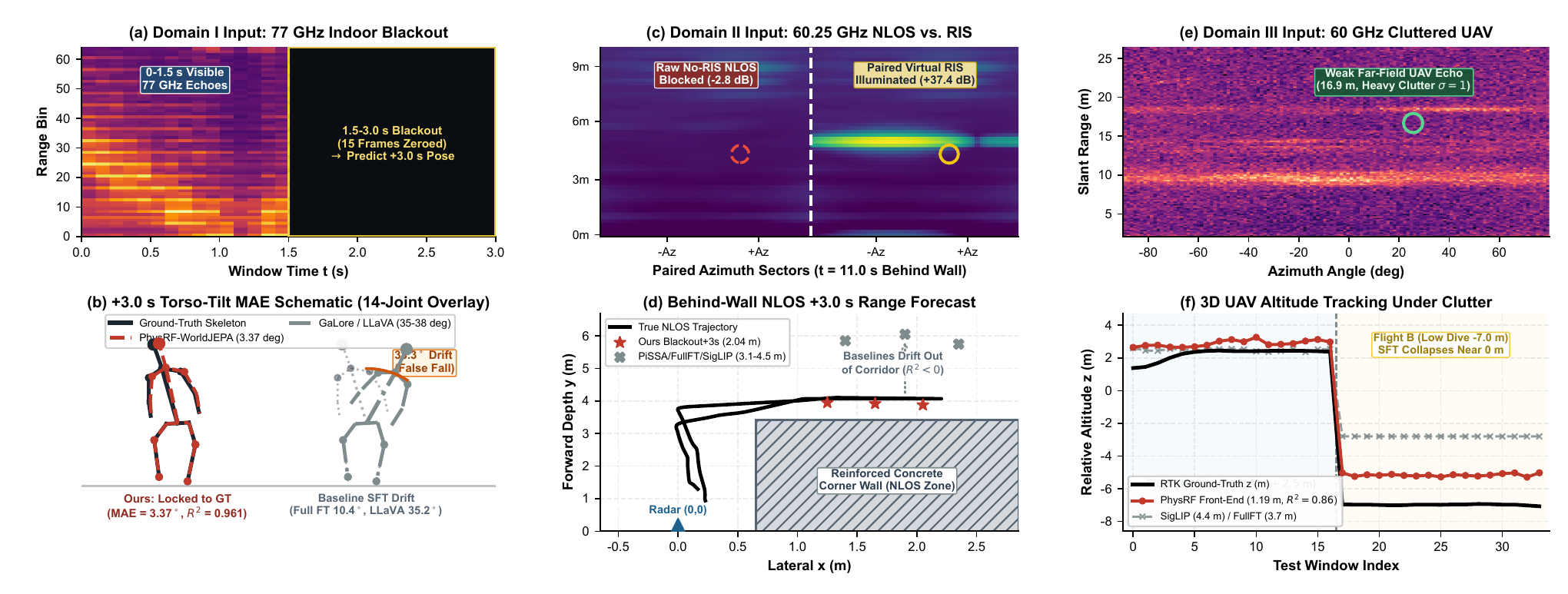}
  \vspace{-0.4em}
  \caption{\textbf{Tri-domain physical state prediction from corrupted continuous mmWave spectra.} \textbf{Top row (a, c, e):} Real RF inputs under severe zero-shot corruptions---(a) $77\,\text{GHz}$ indoor range--time radargram with the trailing $1.5\,\text{s}$ zeroed out; (c) $60.25\,\text{GHz}$ through-wall L-corridor range--azimuth spectrum comparing wall-attenuated raw NLOS ($-2.8\,\text{dB}$) against paired virtual RIS illumination ($+37.4\,\text{dB}$); and (e) $60\,\text{GHz}$ far-field outdoor UAV spectrum buried in $\sigma=1.0$ clutter. \textbf{Bottom row (b, d, f):} (b) \emph{Schematic overlay} illustrating locked-test $+3.0\,\text{s}$ torso-tilt MAE from LLM causal state $\mathbf{h}_t$ ($3.37^\circ$ vs.\ $38.30^\circ$ in GaLore~\cite{zhao2024galore} and $35.15^\circ$ in LLaVA+NEFTune~\cite{liu2024llava15,jain2024neftune}) by rotating a ground-truth 14-joint skeleton around the pelvis (not per-joint keypoint regression); (d) behind-wall NLOS $+3.0\,\text{s}$ range readout from $\mathbf{h}_t$ ($2.04\,\text{m}$ vs.\ $3.1\text{--}4.5\,\text{m}$ drift in PiSSA~\cite{meng2024pissa}/Full FT/SigLIP~\cite{zhai2023siglip}); and (f) cluttered 3D UAV altitude readout from the Stage~1 \texttt{PhysRF} front-end pooled state $\bar{\mathbf{x}}_{\text{rf}}$ ($1.19\,\text{m}$, $R^2\!=\!+0.863$; LLM state $\mathbf{h}_t$ achieves $1.89\,\text{m}$, $R^2\!=\!+0.679$).}
  \label{fig:teaser}
  \vspace{-0.5em}
\end{figure*}

\section{Introduction}
\label{sec:intro}

Millimeter-wave (mmWave) frequency-modulated continuous-wave (FMCW) radar operates at $60\text{--}77\,\text{GHz}$ wavelengths that penetrate smoke, steam, glare, and non-metallic partitions without capturing optically identifiable facial or clothing imagery~\cite{zhao2018rfpose,chen2022mmbody,wu2021ris}. Connecting mmWave radar to pretrained Large Language Models (LLMs) offers an attractive blueprint for unified open-vocabulary physical reasoning across indoor healthcare, through-wall non-line-of-sight (NLOS) security, and low-altitude unmanned aerial vehicle (UAV) traffic management~\cite{lai2025radarllm}.

Yet bridging continuous electromagnetic wave returns and discrete autoregressive language models exposes three structural failure modes that do not arise in standard vision--language modeling:
\begin{enumerate}[leftmargin=1.4em,itemsep=2pt,topsep=2pt]
  \item \textbf{Quantization Loss \& Universal SFT Retrieval Blindness.} Prior radar--LLM work such as RadarLLM~\cite{lai2025radarllm} discretizes mmWave point-cloud sequences via a 512-entry VQ-VAE codebook; when discrete tokenization is adapted to continuous radar spectra in our controlled 64-codebook baseline, sub-wavelength micro-Doppler cues are lost and fall sequence recall drops to $40\%$ ($60\%$ missed). Meanwhile, single-stage multimodal instruction tuning (\texttt{Masked-SFT + LoRA}~\cite{hu2022lora}) confines causal EOS states to a narrow anisotropic cone~\cite{ethayarajh2019contextual,liang2022mindgap}, leaving \emph{every} evaluated LLM family at near-random global retrieval ($\text{MedRank} \in [336, 366]$ vs.\ chance $353.5$).
  \item \textbf{Autoregressive Template Memorization Under Sensor Blackouts.} Modern adaptation recipes---including 1.72B full fine-tuning, DoRA~\cite{liu2024dora}, PiSSA~\cite{meng2024pissa}, GaLore~\cite{zhao2024galore}, and LLaVA+NEFTune~\cite{liu2024llava15,jain2024neftune}---optimize next-token cross-entropy over text templates. When transient multipath fading or corner occlusion blanks out $1.5\,\text{s}$ of radar frames (Fig.~\ref{fig:teaser}a--b), LLaVA+NEFTune and GaLore achieve $>97\%$ clean classification yet drift by $35.15^\circ$ and $38.30^\circ$ ($R^2 < 0$) when forecasting $+3.0\,\text{s}$ human torso tilt.
  \item \textbf{The Geometry--Semantics Seesaw.} Latent predictive architectures (JEPA)~\cite{lecun2022path,assran2023ijepa,bardes2024vjepa} anchor continuous physical dynamics, whereas coupling pairwise contrastive objectives such as SigLIP~\cite{zhai2023siglip} directly into the shared backbone improves median rank to $81\text{--}83$ at the cost of $O(B^2)$ uniform angular repulsion that tears low-sample 3D UAV altitude ($R^2 = -1.15$) and behind-wall NLOS range ($R^2 = -1.98$) manifolds apart.
\end{enumerate}

\begin{figure*}[t]
  \centering
  \includegraphics[width=\textwidth]{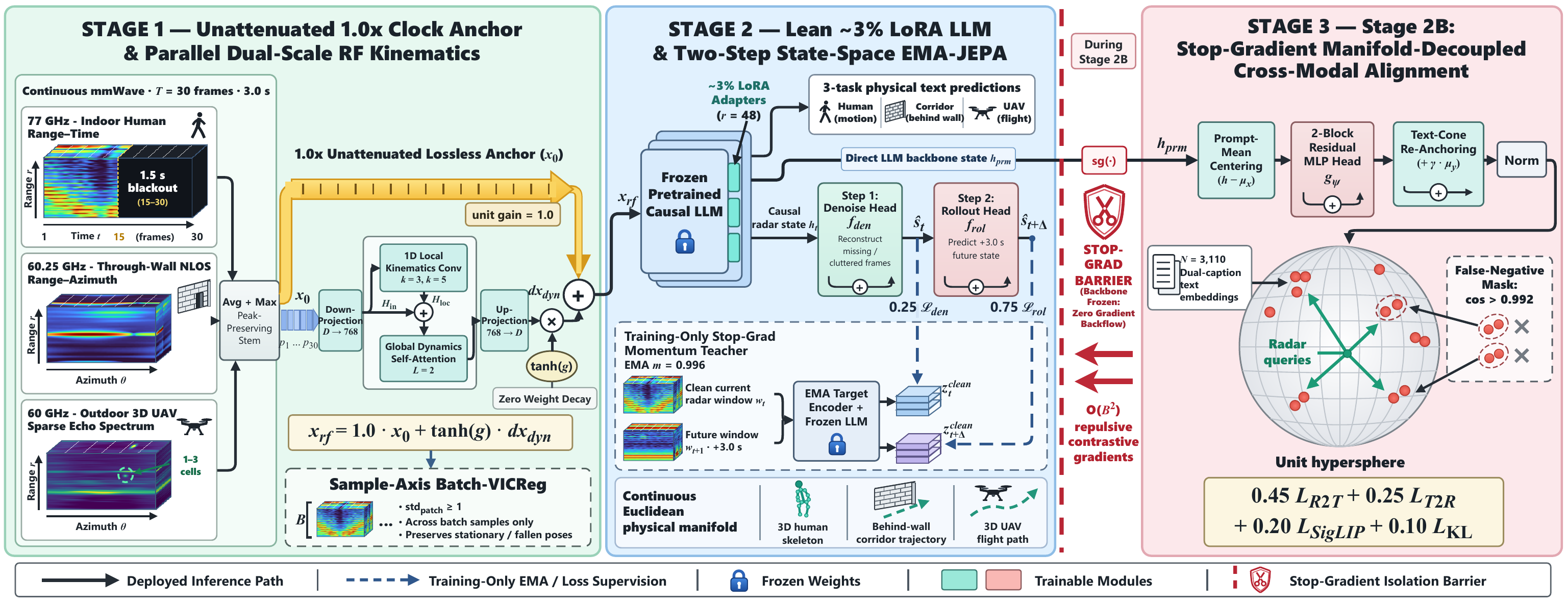}
  \vspace{-0.4em}
  \caption{\textbf{Architecture of PhysRF-WorldJEPA.} \textbf{Stage 1:} Continuous tri-domain RF spectra pass through peak-preserving reduction into an \emph{unattenuated $1.0\times\mathbf{x}_0$ clock anchor} plus a dual-scale RF kinematics residual ($\text{LocalConv} + \text{GlobalAttn}$) gated by zero-weight-decay $\tanh(\mathbf{g})$ with sample-axis Batch-VICReg. \textbf{Stage 2:} A frozen causal LLM with $\sim\!3\%$ LoRA ($r=48$) processes $\mathbf{x}_{\text{rf}}$ jointly with a Two-Step State-Space Predictor ($\hat{\mathbf{s}}_t = f_{\text{den}}(\mathbf{h}_t)$, $\hat{\mathbf{s}}_{t+\Delta} = f_{\text{rol}}(\hat{\mathbf{s}}_t)$) supervised by an EMA teacher ($m=0.996$). \textbf{Stage 3 (Stage 2B):} A \texttt{STOP-GRAD} barrier ($\operatorname{sg}(\cdot)$) blocks $O(B^2)$ repulsive contrastive gradients while residual head $g_\psi$ aligns prompt states $\mathbf{h}_{\text{prm}}$ onto the unit hypersphere with false-negative masking ($\cos > 0.992$).}
  \label{fig:arch}
  \vspace{-0.5em}
\end{figure*}

To resolve all three bottlenecks and establish whether arbitrary local LLMs can serve as electromagnetic world models, we propose \textbf{PhysRF-WorldJEPA} (Fig.~\ref{fig:arch}). Rather than inflating LLM adapter rank---which triggers an overfitting collapse ($19.02^\circ$ blackout error at $5\%$ LoRA)---we pair a lean $\sim\!3\%$ LoRA LLM with a parallel physics-informed RF front-end, a two-step latent state-space predictor, and a gradient-isolated cross-modal alignment head, and further dissect pure zero-overhead objective scaling across five heterogeneous LLM families.

Our main contributions are summarized as follows:
\begin{itemize}[leftmargin=1.3em,itemsep=1.8pt,topsep=2pt]
  \item \textbf{Unattenuated Anchor \& Parallel Dual-Scale RF Kinematics.} We design a lossless tri-domain peak-preserving RF encoder uniting an \emph{unattenuated $1.0\times \mathbf{x}_0$ clock anchor} with parallel local convolutional kinematics ($k\in\{3,5\}$) and global self-attention dynamics, regularized strictly along the batch axis (\emph{Batch-VICReg}) so stationary and fallen postures are never distorted.
  \item \textbf{Two-Step State-Space EMA-JEPA \& Euclidean Norm Calibration.} We factorize latent world prediction into observation denoising followed by $+3.0\,\text{s}$ state-space rollout against a momentum EMA teacher, augmented with dimensionless scale-normalized Euclidean state anchoring ($\operatorname{SmoothL1}(\mathbf{h}_t^{\text{corr}}/\sigma_z, \mathbf{z}_t^{\text{clean}}/\sigma_z)$) to eliminate terminal \texttt{[MASK]} norm drift across instruction and reasoning LLMs.
  \item \textbf{Manifold-Decoupled Cross-Modal Alignment.} We characterize the \emph{Geometry--Semantics Seesaw} and resolve it via a stop-gradient, prompt-centered residual alignment head trained with false-negative-masked ($\cos > 0.992$) InfoNCE, SigLIP, and relational KL distillation---raising $\text{R}\!\rightarrow\!\text{T}$ R@10 from $5.95\%$ to $\mathbf{15.01\%}$ ($\text{MedRank}: 107.0 \to \mathbf{60.0}$) on the frozen physical backbone with zero readout degradation.
  \item \textbf{Comprehensive Tri-Domain Evaluation Against 10 Baselines.} Across 4,541 windows spanning $77\,\text{GHz}$ indoor pose, $60.25\,\text{GHz}$ through-wall NLOS/RIS, and $60\,\text{GHz}$ outdoor 3D RTK UAVs, PhysRF-WorldJEPA is the only model combining strong global retrieval ($\text{MedRank}=60.0$, $\text{R@10}=15.01\%$) with substantial positive tri-domain physical probe $R^2$ ($+0.961, +0.679$ [$+0.863$ at the RF front-end], $+0.461$), $0.9976$ virtual-RIS cosine invariance, $+36.6\%$ low-label blackout macro-recall gain, and $92.9\%$ unseen-subject accuracy.
  \item \textbf{Cross-Family Zero-Overhead Portability Across 5 LLMs (15 Models).} Stripping away custom front-end layers under a strict zero-overhead contract across \texttt{Qwen2.5-0.5B}~\cite{yang2024qwen25}, \texttt{Llama-3.2-1B}~\cite{dubey2024llama3}, \texttt{DeepSeek-R1-Distill-1.5B}~\cite{guo2025deepseekr1}, \texttt{SmolLM2-1.7B}~\cite{allal2025smollm2}, and \texttt{Qwen3-1.7B}~\cite{yang2025qwen3} yields $10/10$ retrieval and $75\%$-drop completion gains (up to $63.5\%$ R@5) and outperforms SFT on blackout $+3.0\,\text{s}$ tilt in $4/5$ families with fixed \texttt{Span-Euc}, $2/5$ with fixed \texttt{Scatter}, and $5/5$ taking the better recipe per family ($5.61^\circ\text{--}13.86^\circ$), with orthogonal gains when adding \texttt{PhysRF} ($3.37^\circ$).
\end{itemize}

\section{Related Work}
\label{sec:related}

\paragraph{Millimeter-Wave Radar Perception and NLOS Sensing.}
Dedicated deep networks recover human skeletons through walls and occlusions~\cite{zhao2018rfpose}, reconstruct 3D human meshes from mmWave heatmaps~\cite{chen2022mmbody}, and exploit reconfigurable intelligent surfaces (RIS) for non-line-of-sight wireless sensing~\cite{wu2021ris}. However, specialist radar architectures are locked to single-sensor tensor shapes and closed-set heads: a converged 4.1M pure-radar specialist reaches $95.0\%$ indoor recall yet stalls at $63.2\%$ on through-wall NLOS reasoning, $17.6\%$ on 3D UAV maneuvers, and lacks open-vocabulary retrieval or forecasting capabilities.

\paragraph{Heterogeneous LLMs and Multimodal Adaptation.}
Visual instruction tuning projects continuous features into LLMs via MLP connectors~\cite{liu2024llava15} with optional noisy embedding regularization (NEFTune~\cite{jain2024neftune}), while parameter-efficient fine-tuning has evolved from LoRA~\cite{hu2022lora} to DoRA~\cite{liu2024dora}, rsLoRA~\cite{kalajdzievski2023rslora}, PiSSA~\cite{meng2024pissa}, LoRA+~\cite{hayou2024loraplus}, GaLore~\cite{zhao2024galore}, and SimPO~\cite{meng2024simpo}. Meanwhile, open-weights edge LLMs span diverse attention topologies and pretraining priors, including sub-billion GQA (\texttt{Qwen2.5-0.5B}~\cite{yang2024qwen25}), 128K TikToken GQA (\texttt{Llama-3.2-1B}~\cite{dubey2024llama3}), 49K full MHA (\texttt{SmolLM2-1.7B}~\cite{allal2025smollm2}), RL-distilled chain-of-thought reasoning (\texttt{DeepSeek-R1-Distill-1.5B}~\cite{guo2025deepseekr1}), and hybrid GQA (\texttt{Qwen3-1.7B}~\cite{yang2025qwen3}). Unlike vision where pretrained CLIP/SigLIP ViTs~\cite{radford2021clip,zhai2023siglip} supply semantically aligned tokens, mmWave radar lacks an internet-scale foundation encoder, causing direct SFT or discrete codebook quantization (e.g., point-cloud VQ-VAE tokenization in RadarLLM~\cite{lai2025radarllm}) to overfit surface templates or lose fine-grained spectral dynamics.

\paragraph{Joint-Embedding Predictive Architectures and Modality Geometry.}
Self-supervised latent prediction---pioneered by CPC~\cite{oord2018cpc}, VICReg~\cite{bardes2022vicreg}, JEPA~\cite{lecun2022path}, I-JEPA~\cite{assran2023ijepa}, V-JEPA~\cite{bardes2024vjepa}, and text-space LLM-JEPA~\cite{huang2025llmjepa}---learns predictable world states without signal-level reconstruction. However, causal LLMs concentrate token states inside a narrow anisotropic cone~\cite{ethayarajh2019contextual} separated by a persistent cross-modal gap~\cite{liang2022mindgap}. Our work bridges continuous electromagnetic JEPA dynamics and heterogeneous LLMs via decoupled gradient pathways and scale-normalized latent anchoring.

\section{Methodology: PhysRF-WorldJEPA}
\label{sec:method}

Given a multi-domain stream of continuous mmWave radar windows $\mathcal{W}_t \in \mathbb{R}^{T \times C_d \times H_d \times W_d}$ spanning $T=30$ frames ($3.0\,\text{s}$ at $10\,\text{Hz}$) in domain $d \in \{\text{Indoor-77GHz},\allowbreak \text{UAV-60GHz},\allowbreak \text{NLOS-60.25GHz}\}$, our goal is to learn a unified radar--LLM world model capable of: (i) accurate physical state estimation under severe RF packet loss and clutter, (ii) zero-shot $+3.0\,\text{s}$ future physical trajectory forecasting ($\mathcal{W}_{t+\Delta}$) across total sensor blackouts, and (iii) fine-grained bidirectional radar--language retrieval and captioning.

\subsection{Lossless Peak-Preserving Tri-Domain RF Tokenization}
\label{sec:tokenizer}

Discrete vector quantization discards subtle phase and amplitude gradients critical for far-field radar targets. Instead, for each domain $d$, we map raw range--azimuth (RA), clutter-mitigated range--azimuth (RA-MIT), and range--Doppler (RD) tensors directly into continuous frame embeddings. Because an outdoor UAV at $15\text{--}25\,\text{m}$ occupies only $1\text{--}3$ range--azimuth cells whose energy is washed out by spatial average pooling, we apply a \emph{dual-stream peak-preserving stem}:
\begin{equation}
  \mathbf{u}_{t,\tau} = \phi_d\!\Big(\big[\,\operatorname{AvgPool}_{2\text{D}}(\mathcal{W}_{t,\tau}) \;\|\; \operatorname{MaxPool}_{2\text{D}}(\mathcal{W}_{t,\tau})\,\big]\Big) \in \mathbb{R}^{D},
\end{equation}
where $D$ matches the LLM hidden dimension ($D=2048$ for \texttt{Qwen3-1.7B}), $\tau \in \{1,\dots,T\}$ indexes frames, and $\phi_d$ is a lightweight 2D stem with LayerNorm. Adding learnable frame-time embeddings $\mathbf{p}_\tau \in \mathbb{R}^D$ and domain embeddings $\mathbf{e}_d \in \mathbb{R}^D$ yields the lossless base token sequence (\texttt{UnifiedRadarEncoder}):
\begin{equation}
  \mathbf{x}_{0,\tau} = \mathbf{u}_{t,\tau} + \mathbf{p}_\tau + \mathbf{e}_d, \qquad \mathbf{X}_0 \in \mathbb{R}^{T \times D}.
  \label{eq:x0}
\end{equation}
Physical RF corruptions---frame-drop masks $m_\tau \in \{0,1\}$ and additive Gaussian clutter $\boldsymbol{\epsilon}_\tau \sim \mathcal{N}(\mathbf{0}, \sigma^2 \operatorname{Var}(\mathcal{W}_t)\mathbf{I})$---are applied \emph{prior} to adding $(\mathbf{p}_\tau + \mathbf{e}_d)$, producing corrupted tokens $\tilde{\mathbf{X}}_0$ whose temporal clock coordinates $\mathbf{p}_\tau$ remain intact even when $\mathcal{W}_{t,\tau}=\mathbf{0}$.

\subsection{Unattenuated $1.0\times$ Anchor and Parallel Dual-Scale RF Kinematics}
\label{sec:physrf}

Expanding LLM LoRA rank from $1\%$ to $5\%$ ($93.7\text{M}$ trainable parameters) causes zero-shot blackout $+3.0\,\text{s}$ tilt error to explode to $19.02^\circ$ due to syntax memorization, whereas an ungated serial RF Transformer suffers rank-1 temporal collapse ($23.5\%$ NLOS accuracy). We resolve both via a \textbf{Parallel Dual-Scale RF Kinematics Encoder} anchored to an \emph{unattenuated $1.0\times$ lossless token highway}. Projecting $\tilde{\mathbf{X}}_0$ to $D_{\text{rf}}=768$, $\mathbf{H}_{\text{in}} = W_{\text{down}}\operatorname{LN}(\tilde{\mathbf{X}}_0) \in \mathbb{R}^{T \times D_{\text{rf}}}$, we compute:
\begin{align}
  \mathbf{H}_{\text{loc}} &= \operatorname{GELU}\big(\operatorname{Conv}_{3}(\mathbf{H}_{\text{in}})\big) + \operatorname{GELU}\big(\operatorname{Conv}_{5}(\mathbf{H}_{\text{in}})\big), \\
  \mathbf{H}_{\text{glo}} &= \operatorname{TransEnc}_{L=2, H=12}\big(\mathbf{H}_{\text{in}} + \mathbf{H}_{\text{loc}}\big),
\end{align}
where $k=3$ ($0.3\,\text{s}$) captures limb swing and radial velocity, $k=5$ ($0.5\,\text{s}$) captures torso acceleration and fall onset, and $L_{\text{rf}}=2$ self-attention layers propagate kinematics across $1.5\,\text{s}$ blackout gaps:
\begin{equation}
  \mathbf{X}_{\text{rf}} = \underbrace{1.0 \cdot \tilde{\mathbf{X}}_0}_{\text{Unattenuated Anchor}} + \tanh(\mathbf{g}) \odot W_{\text{up}}\operatorname{LN}(\mathbf{H}_{\text{glo}}),
  \label{eq:anchor}
\end{equation}
with $\mathbf{g} \in \mathbb{R}^D$ initialized to $0.20$ under \emph{zero weight decay}. Unlike convex Kalman gating $\mathbf{w}_{\text{obs}} \odot \tilde{\mathbf{X}}_0 + (1-\mathbf{w}_{\text{obs}})\odot \hat{\mathbf{X}}_{\text{dyn}}$ ($\mathbf{w}_{\text{obs}} \in (0,1)^D$), which dampens terminal clock indices $\mathbf{p}_{16:30}$ during blackouts and doubles forecast error ($3.37^\circ \to 7.13^\circ$), unit gain $1.0\cdot\tilde{\mathbf{X}}_0$ preserves an exact temporal coordinate ruler.

\paragraph{Sample-Axis Batch-VICReg.}
To prevent dimensional collapse without distorting stationary physics, hinge variance regularization is applied \emph{strictly across batch axis $B$} on $\bar{\mathbf{x}}_b = \frac{1}{T}\sum_{\tau=1}^T \mathbf{X}_{\text{rf},b,\tau}$:
\begin{equation}
  \mathcal{L}_{\text{var}}(\mathbf{X}_{\text{rf}}) = \frac{1}{D}\sum_{k=1}^{D} \max\!\left(0,\; 1 - \sqrt{\operatorname{Var}_{b=1..B}(\bar{x}_{b,k}) + \varepsilon}\right),
  \label{eq:vicreg}
\end{equation}
deliberately omitting temporal-axis variance penalties ($\operatorname{Var}_\tau$) that penalize upright standing, lying after a fall, or UAV hovering ($100\% \to 40\%$ fall recall).

\begin{figure*}[t]
  \centering
  \includegraphics[width=\textwidth]{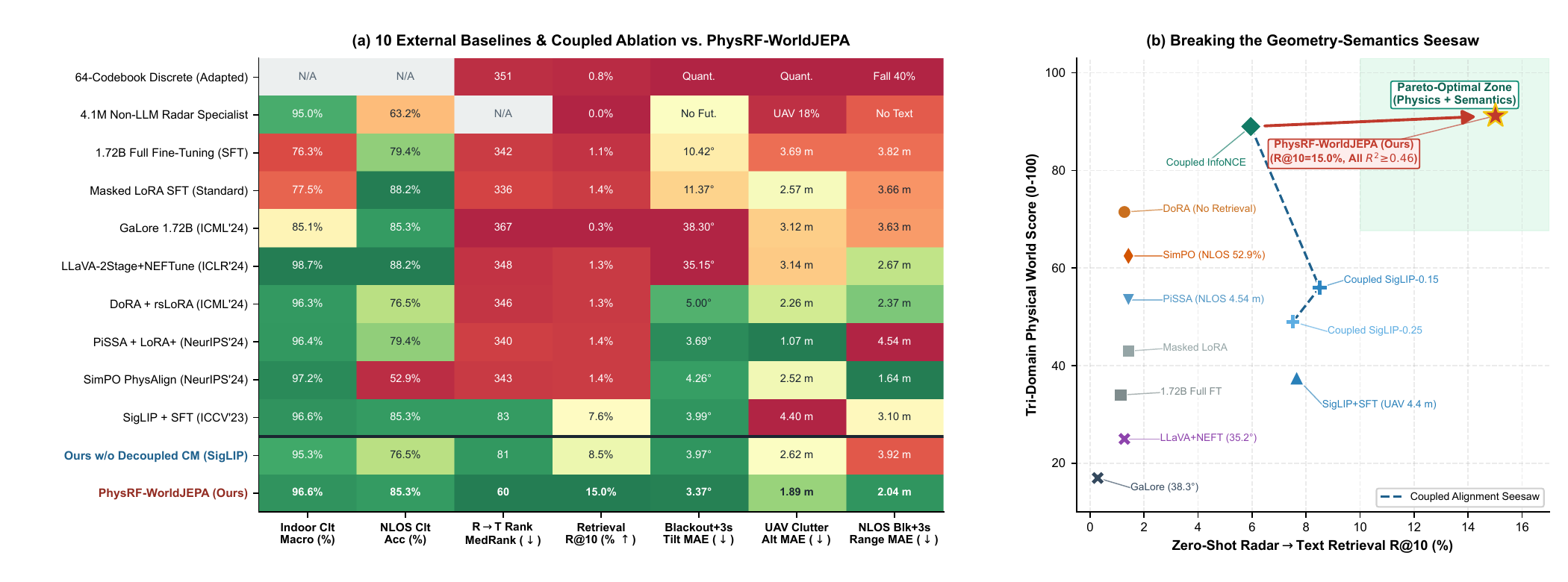}
  \vspace{-0.4em}
  \caption{\textbf{Comprehensive benchmark against 10 external baselines and resolution of the Geometry--Semantics Seesaw.} \textbf{(a)} Normalized capability matrix across 7 key metrics spanning four external baseline families: (1) discrete-codebook RadarLLM~\cite{lai2025radarllm} and a 4.1M pure-radar specialist, (2) full-parameter SFT and GaLore~\cite{zhao2024galore}, (3) multimodal SFT (LLaVA+NEFTune~\cite{liu2024llava15,jain2024neftune}) and modern PEFT/alignment (DoRA+rsLoRA~\cite{liu2024dora,kalajdzievski2023rslora}, PiSSA+LoRA+~\cite{meng2024pissa,hayou2024loraplus}, SimPO~\cite{meng2024simpo}), and (4) discriminative contrastive SigLIP+SFT~\cite{zhai2023siglip}. \textbf{(b)} Coupled alignment methods trace a strict seesaw between retrieval R@10 and tri-domain physical accuracy; Stage~2B manifold decoupling (\textbf{PhysRF-WorldJEPA}) breaks directly into the upper-right Pareto-optimal zone.}
  \label{fig:matrix}
  \vspace{-0.5em}
\end{figure*}

\subsection{Two-Step State-Space EMA-JEPA ($\text{Denoise} \rightarrow \text{Rollout}$)}
\label{sec:twostep}

Let $f_\theta$ denote the causal LLM with $\sim\!3\%$ LoRA adapters on $\{q,k,v,o,\text{gate},\text{up},\text{down}\}$. Passing $\mathbf{X}_{\text{rf}}$ through $f_\theta$ without text tokens yields causal radar state $\mathbf{h}_t^{\text{corr}} = f_\theta(\mathbf{X}_{\text{rf}})_{[:,-1]} \in \mathbb{R}^D$, while a stop-gradient EMA target encoder ($m=0.996$) encodes clean windows $\mathcal{W}_t$ and $\mathcal{W}_{t+\Delta}$ ($\Delta=+3.0\,\text{s}$) into $\mathbf{z}_t^{\text{clean}}$ and $\mathbf{z}_{t+\Delta}^{\text{clean}}$. To prevent clutter removal and temporal extrapolation from competing inside one MLP, we factorize prediction into a \textbf{Two-Step State-Space Predictor}:
\begin{equation}
  \hat{\mathbf{s}}_t = \mathbf{h}_t^{\text{corr}} + \psi_{\text{den}}(\mathbf{h}_t^{\text{corr}}), \qquad
  \hat{\mathbf{s}}_{t+\Delta} = \hat{\mathbf{s}}_t + \psi_{\text{rol}}(\hat{\mathbf{s}}_t).
  \label{eq:twostep}
\end{equation}
Using cosine InfoNCE $\mathcal{L}_{\text{nce}}(\mathbf{u}, \mathbf{v}; \tau_p=0.10)$, Stage~1 pretrains label-free dynamics ($0.75\,\mathcal{L}_{\text{nce}}(\hat{\mathbf{s}}_{t+\Delta}, \mathbf{z}_{t+\Delta}^{\text{clean}}) + 0.25\,\mathcal{L}_{\text{nce}}(\hat{\mathbf{s}}_t, \mathbf{z}_t^{\text{clean}}) + 0.02\,\mathcal{L}_{\text{var}}$ with LoRA dropout $p=0$), followed by Stage~2A joint three-task instruction tuning ($p=0.05$) with prompt-pure caption InfoNCE:
\begin{equation}
  \mathcal{L}_{\text{Stage2A}} = \mathcal{L}_{\text{CE}} + 0.22\,\mathcal{L}_{\text{cm}}^{\text{gentle}} + 0.06\,\mathcal{L}_{\text{rol}} + 0.03\,\mathcal{L}_{\text{den}} + 0.01\,\mathcal{L}_{\text{var}}.
  \label{eq:stage2a}
\end{equation}

\subsection{Gradient-Conflict Analysis and Manifold-Decoupled Alignment}
\label{sec:decoupled}

When pairwise sigmoid contrastive loss (SigLIP~\cite{zhai2023siglip}, temperature $t_s \gg 1$) is coupled directly into $\mathbf{h}_i$, every $j \neq i$ in batch $B$---including physically adjacent states $|y_i - y_j| < 0.2\,\text{m}$---injects repulsive gradient $\frac{t_s}{B}\sigma(t_s\langle \hat{\mathbf{h}}_i, \hat{\mathbf{y}}_j\rangle + b_s)\hat{\mathbf{y}}_j$, shattering minority UAV and NLOS coordinate smoothness ($R^2 \to -1.982$, Fig.~\ref{fig:matrix}b). Stage~2B resolves this \emph{Geometry--Semantics Seesaw} via a strict \texttt{stop-grad} barrier ($\nabla_\theta = \mathbf{0}, \nabla_{\phi_{\text{rf}}} = \mathbf{0}$), training a prompt-centered residual head $g_\psi$ on frozen states $\mathbf{h}_i = \operatorname{sg}(\mathbf{h}_i^{\text{cap}})$:
\begin{align}
  \tilde{\mathbf{h}}_i &= \mathbf{h}_i - \boldsymbol{\mu}_x, \quad
  \mathbf{u}_i = W_{\text{base}}\operatorname{LN}(\tilde{\mathbf{h}}_i) + \mathcal{B}_1(\tilde{\mathbf{h}}_i) + \mathcal{B}_2(\tilde{\mathbf{h}}_i), \nonumber\\
  \mathbf{z}_i^{\text{rad}} &= \operatorname{Norm}\!\big(W_{\text{out}}\operatorname{LN}(\mathbf{u}_i) + \gamma \boldsymbol{\mu}_y\big),
  \label{eq:cmhead}
\end{align}
where $\boldsymbol{\mu}_x = \mathbb{E}[\mathbf{h}_i]$ removes shared prompt bias and $\gamma\boldsymbol{\mu}_y$ re-anchors vectors into the dual-caption text cone. Masking paraphrase collisions across all $N=3{,}110$ cached training captions via $M_{ij} = \mathbb{I}[\langle \mathbf{y}_i, \mathbf{y}_j \rangle > 0.992 \wedge j \neq i]$, Stage~2B optimizes:
\begin{equation}
  \begin{aligned}
  \mathcal{L}_{\text{Stage2B}} &= 0.45\,\mathcal{L}_{\text{R2T}}^{\text{mask}} + 0.25\,\mathcal{L}_{\text{T2R}} + 0.20\,\mathcal{L}_{\text{SigLIP}}^{\text{mask}} \\
  &\quad + 0.10\,D_{\text{KL}}\!\Big(p_{\tau_k}(\mathbf{z}_i^{\text{rad}}, \mathbf{Y}^{(0)}) \,\Big\|\, q_{\tau_k}(\mathbf{y}_i^{(0)}, \mathbf{Y}^{(0)})\Big).
  \end{aligned}
  \label{eq:stage2b}
\end{equation}

\begin{table*}[t]
  \centering
  \caption{\textbf{Main comparison of semantics, global cross-modal retrieval, and zero-shot degraded perception across 706 locked test windows on \texttt{Qwen3-1.7B}.} \textbf{Fall 8/8} counts how many of the 8 degradation modes maintain $100\%$ fall sequence recall (minimum recall in parentheses). Best results in \textcolor{darkred}{\textbf{bold red}}, second best \underline{underlined}.}
  \label{tab:main1}
  \footnotesize
  \setlength{\tabcolsep}{4.1pt}
  \renewcommand{\arraystretch}{1.07}
  \resizebox{\textwidth}{!}{\begin{tabular}{l r r c c c c c c c c c}
    \toprule
    \multirow{2}{*}{\textbf{Method / Paradigm}} & \multirow{2}{*}{\textbf{Train P.}} & \multirow{2}{*}{\textbf{VRAM}} & \textbf{Cap} & \multicolumn{3}{c}{\textbf{706-Win $\text{Radar}\!\rightarrow\!\text{Text}$ Retrieval}} & \multicolumn{3}{c}{\textbf{Indoor Human Action (\%)}} & \multicolumn{2}{c}{\textbf{Through-Wall NLOS (\%)}} \\
    \cmidrule(lr){5-7} \cmidrule(lr){8-10} \cmidrule(lr){11-12}
    & & & \textbf{NLL}$\downarrow$ & \textbf{R@5}$\uparrow$ & \textbf{R@10}$\uparrow$ & \textbf{MedRk}$\downarrow$ & \textbf{Clean}$\uparrow$ & \textbf{Drop75}$\uparrow$ & \textbf{Clutter}$\uparrow$ / \textbf{Fall 8/8} & \textbf{Blackout}$\uparrow$ & \textbf{Clutter}$\uparrow$ \\
    \midrule
    \multicolumn{12}{l}{\textit{Family I: Discrete Tokenization \& Non-LLM Radar Specialists}} \\
    64-Codebook Discrete (RadarLLM-style)$^*$~\cite{lai2025radarllm} & 1,720M & 15.2G & --- & 0.57 & 0.85 & 351.0 & 74.5 & --- & --- \;(0/8, 40\%) & --- & --- \\
    4.1M Non-LLM Radar Specialist & 4.1M & 1.4G & --- & --- & --- & --- & 94.2 & 93.5 & 95.0 \;(8/8, 100\%) & 66.2 & 63.2 \\
    Projector-Only (Frozen LLM) & 6.5M & 8.7G & 1.518 & 0.57 & 0.99 & 351.0 & 74.0 & 39.8 & 71.2 \;(4/8, 60\%) & 44.1 & 38.2 \\
    \midrule
    \multicolumn{12}{l}{\textit{Family II: Full-Parameter Supervised Fine-Tuning \& Gradient Projection}} \\
    1.72B Full Fine-Tuning (SFT) & 1,726M & 15.3G & 0.261 & 0.57 & 1.13 & 342.0 & 76.0 & 75.6 & 76.3 \;(0/8, 40\%) & \secbest{86.8} & 79.4 \\
    GaLore 1.72B~\cite{zhao2024galore} (ICML'24) & 1,726M & 13.9G & 0.330 & 0.28 & 0.28 & 367.0 & 97.2 & 60.8 & 85.1 \;(5/8, 20\%) & 32.4 & \secbest{85.3} \\
    \midrule
    \multicolumn{12}{l}{\textit{Family III: Modern Multimodal Instruction Tuning, PEFT, \& Preference Alignment}} \\
    Masked LoRA SFT~\cite{hu2022lora} & 23.9M & 6.1G & 0.249 & 0.71 & 1.42 & 336.0 & 88.8 & 76.5 & 77.5 \;(0/8, 60\%) & \best{88.2} & \best{88.2} \\
    LLaVA-2Stage + NEFTune~\cite{liu2024llava15,jain2024neftune} & 22.8M & 8.8G & \best{0.234} & 0.71 & 1.27 & 348.0 & \best{98.2} & \best{97.4} & \best{98.7} \;(8/8, 100\%) & 69.1 & \best{88.2} \\
    DoRA + rsLoRA~\cite{liu2024dora,kalajdzievski2023rslora} (ICML'24) & 58.2M & 6.5G & 0.246 & 0.71 & 1.27 & 346.0 & 96.0 & 95.2 & 96.3 \;(8/8, 100\%) & 83.8 & 76.5 \\
    PiSSA + LoRA+~\cite{meng2024pissa,hayou2024loraplus} (NeurIPS'24) & 22.8M & 6.1G & 0.249 & 0.71 & 1.42 & 340.0 & 96.3 & 96.0 & 96.4 \;(8/8, 100\%) & 83.8 & 79.4 \\
    SimPO PhysAlign~\cite{meng2024simpo} (NeurIPS'24) & 22.8M & 6.9G & 0.256 & 0.71 & 1.42 & 343.0 & 96.9 & \secbest{97.2} & \secbest{97.2} \;(8/8, 100\%) & 79.4 & 52.9 \\
    \midrule
    \multicolumn{12}{l}{\textit{Family IV: Coupled Cross-Modal Contrastive Alignment}} \\
    SigLIP + SFT~\cite{zhai2023siglip} (ICCV'23) & 24.9M & 6.1G & \secbest{0.243} & 3.97 & 7.65 & 83.0 & \secbest{96.9} & 95.9 & 96.6 \;(8/8, 100\%) & 83.8 & \secbest{85.3} \\
    \midrule
    \multicolumn{12}{l}{\textit{Predictive World Modeling \& Manifold-Decoupled Alignment (Ours)}} \\
    \rowcolor{litebg}
    PhysRF-WorldJEPA-Lite (1\% LLM) & 32.9M & 6.3G & 0.257 & \secbest{4.25} & \secbest{8.36} & \secbest{79.0} & 96.3 & 95.3 & 96.5 \;(8/8, 100\%) & 80.9 & 76.5 \\
    \rowcolor{bestbg}
    \textbf{PhysRF-WorldJEPA (Ours, 3\% LLM)} & 83.1M$^\dagger$ & 7.6G & 0.253 & \best{9.77} & \best{15.01} & \best{60.0} & 94.8 & 94.0 & 96.6 \;\textbf{(8/8, 100\%)} & \secbest{86.8} & \secbest{85.3} \\
    \bottomrule
  \end{tabular}}
  \vspace{1pt}
  \begin{flushleft}
    \scriptsize $^*$Our controlled 64-codebook spectrum-discretization adaptation inspired by RadarLLM~\cite{lai2025radarllm} (whose original system operates on mmWave point-cloud sequences with a 512-entry VQ-VAE codebook). $^\dagger$Core backbone: $24.5\text{M}$ (PhysRF encoder) $+ 52.3\text{M}$ ($3.04\%$ LoRA) $+ 6.3\text{M}$ (two-step predictor) $= 83.1\text{M}$ ($7.64\,\text{GiB}$ VRAM); Stage~2B adds a standalone $18.9\text{M}$ decoupled alignment head trained on cached features in $125.8\,\text{s}$. Outdoor 6-class UAV maneuver accuracy reaches $20.6\%$ vs.\ $13.2\%$ on 1.72B Full FT and LLaVA+NEFTune.
  \end{flushleft}
  \vspace{-0.5em}
\end{table*}

\subsection{Zero-Overhead Cross-LLM Scaling \& Euclidean Calibration}
\label{sec:crossllm_method}

To test whether arbitrary local LLMs $\mathcal{M}_k = (f_{\theta_k}, \mathcal{T}_k, D_k)$ ($D_k \in \{896, 1536, 2048\}$) become physical world models \emph{purely} through our JEPA objective---without \texttt{PhysRF} layers or the decoupled alignment head---we enforce a strict \textbf{zero-overhead iso-parameter contract} directly on $\tilde{\mathbf{X}}_0$ (\texttt{UnifiedRadarEncoder}): width-scaled LoRA $(r_k, \alpha_k) \in \{(24,48), (36,64), (48,64)\}$ holding trainable LLM fraction at $\sim\!2.3\%\text{--}3.2\%$, right-padded EOS indexing across all $49\text{K}\text{--}151\text{K}$ tokenizers, and EMA teacher discard at deployment ($0$ extra inference parameters). Beyond scattered-drop rollout (\texttt{JEPA-Scatter}), we observe that instruction-tuned and CoT-distilled LLMs allocate high-norm residual sinks to terminal \texttt{[MASK]} tokens; because $\ell_2$-normalized InfoNCE is blind to vector norm, contiguous trailing blackouts induce affine probe shifts. We therefore formulate \texttt{JEPA-Span-Euc}, combining gentle stem-first Stage~1 warmup ($\text{lr}_{\text{llm}}=2\times 10^{-5}$), contiguous $8\text{--}16$ frame span masking, prompt-pure $\mathcal{L}_{\text{cm}}$, and a \textbf{dimensionless scale-normalized Euclidean state anchor}:
\begin{equation}
  \begin{aligned}
    \mathcal{L}_{\text{self}} ={}& \lambda_{\text{nce}}\,\mathcal{L}_{\text{nce}}(\mathbf{h}_t^{\text{corr}}, \mathbf{z}_t^{\text{clean}}) \\
    &+ \lambda_{\text{euc}}\,\operatorname{SmoothL1}\!\big(\mathbf{h}_t^{\text{corr}}/\sigma_z,\, \mathbf{z}_t^{\text{clean}}/\sigma_z\big),
  \end{aligned}
  \label{eq:spaneuc}
\end{equation}
where $\sigma_z = \operatorname{sg}(\operatorname{std}(\mathbf{z}_t^{\text{clean}}))$ unifies hidden-state RMS scales across heterogeneous LLM families.

\begin{figure*}[t]
  \centering
  \includegraphics[width=\textwidth]{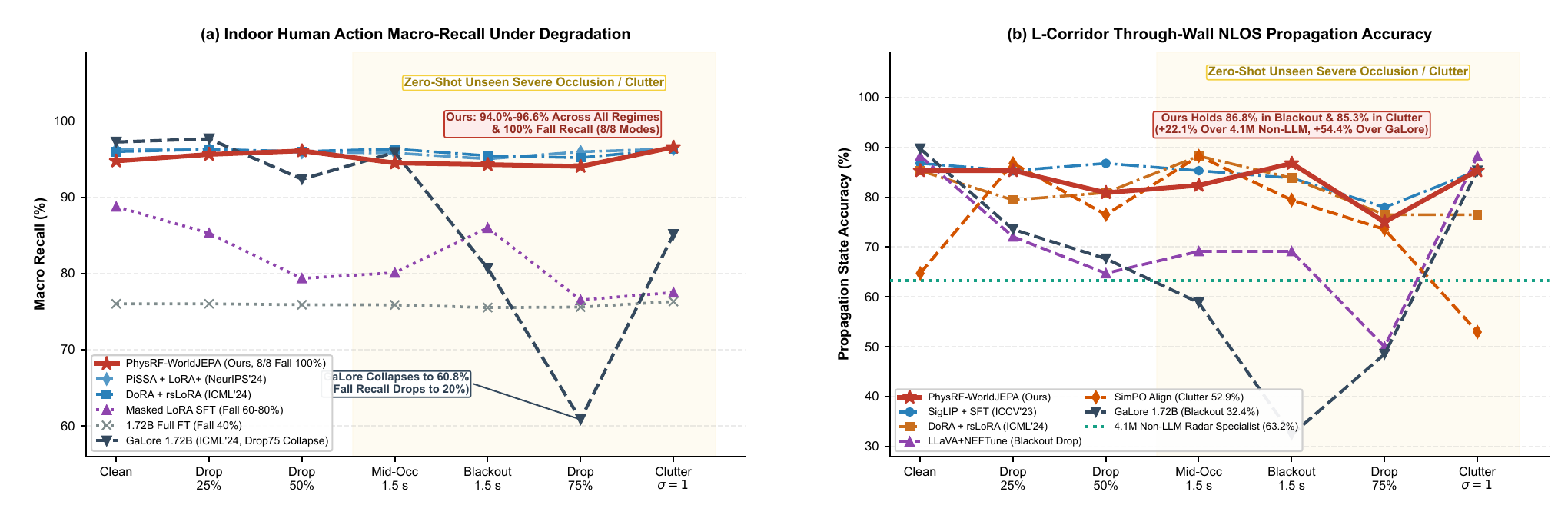}
  \vspace{-0.4em}
  \caption{\textbf{Zero-shot robustness across 7 progressively severe RF degradation regimes.} Models are trained with $0\text{--}50\%$ frame masking and moderate clutter, then tested zero-shot on contiguous $1.5\,\text{s}$ mid-occlusion, $1.5\,\text{s}$ trailing blackout, $75\%$ frame drop, and $\sigma=1.0$ heavy clutter across \textbf{(a)} indoor action/fall recall and \textbf{(b)} through-wall L-corridor NLOS accuracy.}
  \label{fig:degradation}
  \vspace{-0.35em}
\end{figure*}

\begin{table*}[t]
  \centering
  \caption{\textbf{Zero-shot physical state readout on \texttt{Qwen3-1.7B}: latent completion, $+3.0\,\text{s}$ rollout, and continuous 3-domain physical coordinate probes.} All rows probe the identical \textbf{LLM terminal causal state $\mathbf{h}_t$} via frozen Ridge regression ($\alpha=100$) fitted strictly on clean training latents; $+3.0\,\text{s}$ targets are ground-truth states of the immediately following non-overlapping $3.0\,\text{s}$ window (frames $30\text{--}59$ after input frames $0\text{--}29$). Negative $R^2$ (gray) indicates collapse worse than predicting the mean. \textbf{PhysRF-WorldJEPA} is the only model combining strong retrieval ($\text{MedRank}=60.0$) with substantial positive $R^2$ across all three domains simultaneously.}
  \label{tab:probes}
  \footnotesize
  \setlength{\tabcolsep}{3.9pt}
  \renewcommand{\arraystretch}{1.07}
  \resizebox{\textwidth}{!}{\begin{tabular}{l c c c c c c c c c}
    \toprule
    \multirow{2}{*}{\textbf{Method}} & \multicolumn{3}{c}{\textbf{Latent Completion \& $+3\text{s}$ R@5 (\%)}} & \multicolumn{2}{c}{\textbf{Indoor Torso Tilt MAE ($^\circ$)$\downarrow$}} & \multicolumn{2}{c}{\textbf{3D UAV Clutter Alt.}} & \multicolumn{2}{c}{\textbf{NLOS Blackout+$3\text{s}$}} \\
    \cmidrule(lr){2-4} \cmidrule(lr){5-6} \cmidrule(lr){7-8} \cmidrule(lr){9-10}
    & \textbf{Drop75}$\uparrow$ & \textbf{Blk$\to$Cln}$\uparrow$ & \textbf{Blk$\to$Fut}$\uparrow$ & \textbf{Clutter}$\downarrow$ & \textbf{Blackout+$3\text{s}$} ($R^2$) & \textbf{MAE ($\text{m}$)}$\downarrow$ & $R^2\uparrow$ & \textbf{MAE ($\text{m}$)}$\downarrow$ & $R^2\uparrow$ \\
    \midrule
    1.72B Full Fine-Tuning & 12.6 & 15.2 & 6.8 & 8.48 & 10.42 \;(0.792) & 3.69 & \textcolor{gray}{$-1.011$} & 3.82 & \textcolor{gray}{$-1.306$} \\
    Masked LoRA SFT~\cite{hu2022lora} & 11.3 & 34.7 & 7.2 & 6.37 & 11.37 \;(0.692) & 2.57 & $+0.011$ & 3.66 & \textcolor{gray}{$-0.890$} \\
    GaLore 1.72B~\cite{zhao2024galore} & 17.1 & 33.4 & 11.9 & 5.53 & 38.30 \;(\textcolor{gray}{$-0.548$}) & 3.12 & \textcolor{gray}{$-0.040$} & 3.63 & \textcolor{gray}{$-0.853$} \\
    LLaVA-2Stage + NEFTune~\cite{liu2024llava15,jain2024neftune} & 23.8 & 22.4 & 8.5 & 2.58 & 35.15 \;(\textcolor{gray}{$-0.265$}) & 3.14 & \textcolor{gray}{$-0.204$} & 2.67 & \textcolor{gray}{$-0.369$} \\
    DoRA + rsLoRA~\cite{liu2024dora,kalajdzievski2023rslora} & 21.2 & 29.9 & 11.2 & 3.40 & 5.00 \;(0.940) & 2.26 & $+0.493$ & 2.37 & \textcolor{gray}{$-0.314$} \\
    PiSSA + LoRA+~\cite{meng2024pissa,hayou2024loraplus} & 19.0 & 16.4 & 5.7 & \secbest{1.99} & 3.69 \;(\secbest{0.978}) & \best{1.07} & \best{+0.849} & 4.54 & \textcolor{gray}{$-1.534$} \\
    SimPO PhysAlign~\cite{meng2024simpo} & 19.0 & \best{38.1} & 11.3 & \best{1.96} & 4.26 \;(0.963) & 2.52 & $+0.090$ & \best{1.64} & \best{+0.466} \\
    SigLIP + SFT~\cite{zhai2023siglip} & 16.6 & 22.8 & 11.2 & 2.53 & 3.99 \;(\best{0.979}) & 4.40 & \textcolor{gray}{$-1.145$} & 3.10 & \textcolor{gray}{$-0.295$} \\
    \midrule
    Ours w/o Decoupled CM (SigLIP-0.15) & \secbest{31.7} & 19.6 & \secbest{13.2} & 2.65 & 3.97 \;(0.910) & 2.62 & $+0.187$ & 3.92 & \textcolor{gray}{$-1.982$} \\
    \rowcolor{litebg}
    PhysRF-WorldJEPA-Lite (32.9M) & \best{33.1} & 24.5 & \best{13.9} & 2.40 & \best{2.79} \;(0.975) & 2.98 & \textcolor{gray}{$-0.315$} & 3.43 & \textcolor{gray}{$-0.701$} \\
    \rowcolor{bestbg}
    \textbf{PhysRF-WorldJEPA (Ours)} & 29.5 & \secbest{25.2} & 12.5 & 2.36 & \secbest{3.37} \;(0.961) & \secbest{1.89}$^\ddagger$ & \secbest{+0.679}$^\ddagger$ & \secbest{2.04} & \secbest{+0.461} \\
    \bottomrule
  \end{tabular}}
  \vspace{1pt}
  \begin{flushleft}
    \scriptsize $^\ddagger$All cells strictly probe the LLM terminal causal state $\mathbf{h}_t$ for apples-to-apples comparison. Reading out directly from the Stage~1 \texttt{PhysRF} front-end pooled state $\bar{\mathbf{x}}_{\text{rf}}$ further improves cluttered 3D UAV altitude to $\mathbf{1.19\,\text{m}}$ ($R^2=\mathbf{+0.863}$) for PhysRF-WorldJEPA ($2.04\,\text{m}$/$+0.559$ for Lite; $2.27\,\text{m}$/$+0.553$ for coupled SigLIP-0.15).
  \end{flushleft}
  \vspace{-0.45em}
\end{table*}

\section{Experimental Setup}
\label{sec:setup}

\paragraph{Tri-Domain Continuous mmWave Evaluation Suite.}
We evaluate all models on a leakage-free tri-domain evaluation suite comprising \textbf{4,541 core synchronized $3.0\,\text{s}$ windows} ($30$ frames at $10\,\text{Hz}$; \textbf{3,110 train / 725 validation / 706 locked test}, $0$ optical/camera vocabulary leakage):
(1) \textbf{Indoor $77\,\text{GHz}$ Human Pose \& Fall Safety (RVHAD~\cite{jiang2025rvhad}):} 3,895 core windows ($2{,}736 / 589 / 570$ across sequence-disjoint splits) with 14-joint torso tilt ($\theta \in [0^\circ,90^\circ]$), $+3.0\,\text{s}$ future states, and \texttt{standing/sitting/falling} labels, plus a separate held-out cross-subject OOD test set of 608 windows from unseen \texttt{Subject~U1} (excluded from the 4,541 core split);
(2) \textbf{Through-Wall $60.25\,\text{GHz}$ L-Corridor NLOS \& Virtual RIS:} 340 in-house windows ($204 / 68 / 68$) across \texttt{los}, \texttt{partial\_occlusion}, and behind-wall \texttt{full\_nlos}, including 17 paired raw-no-RIS vs.\ $+37.4\,\text{dB}$ virtual-RIS test trajectories;
and (3) \textbf{Outdoor $60\,\text{GHz}$ 3D RTK UAV Flight~\cite{jiang2026uav}:} 306 windows ($170 / 68 / 68$) with centimeter-level RTK-GPS 3D altitude/slant-range coordinates and 6-class maneuvers.
By construction, $3{,}895 + 340 + 306 = 4{,}541$ ($2{,}736+204+170=3{,}110$ train, $589+68+68=725$ val, $570+68+68=706$ test). All splits are strictly sequence-disjoint across train/val/test ($0$ shared recordings): $144/31/30$ indoor sequences ($3.0\,\text{s}$ non-overlapping stride), $9/3/3$ physical L-corridor captures ($12/4/4$ tracks with paired raw-no-RIS and virtual-RIS views locked to the same split, $1.5\,\text{s}$ within-split stride), and $10/4/4$ independent UAV flights ($1.5\,\text{s}$ within-split stride).

\paragraph{Baselines, LLM Zoo, and Five-Dimensional Evaluation.}
On \texttt{Qwen3-1.7B}~\cite{yang2025qwen3}, we benchmark against \textbf{10 modern baselines} under identical $p_d \propto N_d^{0.5}$ sampling, RF augmentations, and 2,400-step validation selection (Table~\ref{tab:main1}). In addition, we train a \textbf{15-model cross-architecture matrix} across five LLM families (\texttt{Qwen2.5-0.5B}, \texttt{Llama-3.2-1B}, \texttt{DeepSeek-R1-Distill-1.5B}, \texttt{SmolLM2-1.7B}, \texttt{Qwen3-1.7B}) comparing \texttt{Masked-SFT}, \texttt{JEPA-Scatter}, and \texttt{JEPA-Span-Euc}. Every model undergoes locked zero-shot testing across: (i) global 706-candidate $\text{R}\rightarrow\text{T}$ retrieval, (ii) 8 RF degradation regimes (\texttt{Clean}, \texttt{Drop25/50/75}, $1.5\,\text{s}$ \texttt{Mid-Occ}, $1.5\,\text{s}$ \texttt{Suffix-Blackout}, $\sigma=1.0$ \texttt{Clutter}, \texttt{Time-Reverse}), (iii) 706-candidate latent completion and $+3.0\,\text{s}$ rollout retrieval, (iv) frozen Ridge physical coordinate probes ($\alpha=100$) fitted strictly on clean training LLM terminal causal states $\mathbf{h}_t$ to predict current or $+3.0\,\text{s}$ future physical states (defined as ground-truth targets of the immediately following non-overlapping $3.0\,\text{s}$ window, frames $30\text{--}59$ after input frames $0\text{--}29$), and (v) paired virtual-RIS cosine invariance.

\begin{table*}[t]
  \centering
  \caption{\textbf{Cross-architecture generalization across five heterogeneous local LLM families (15 zero-overhead models + full \texttt{PhysRF} bridge, 706 test windows).} Rows 1--15 share the \emph{identical} minimal \texttt{UnifiedRadarEncoder} stem and width-scaled $\sim\!3\%$ LoRA ($0$ extra inference parameters). Best zero-overhead metric per family in \textcolor{darkred}{\textbf{bold red}}; bottom row shows orthogonal amplification when adding \texttt{PhysRF} + Stage~2B on \texttt{Qwen3-1.7B}.}
  \label{tab:cross_llm}
  \footnotesize
  \setlength{\tabcolsep}{3.7pt}
  \renewcommand{\arraystretch}{1.05}
  \resizebox{\textwidth}{!}{\begin{tabular}{l c l c c c c c c c c c c}
    \toprule
    \multirow{2}{*}{\textbf{Local LLM Family}} & \multirow{2}{*}{$D$ / \textbf{LoRA}} & \multirow{2}{*}{\textbf{Iso-Param Paradigm}} & \multicolumn{3}{c}{\textbf{706-Win $\text{R}\!\rightarrow\!\text{T}$ Retrieval}} & \textbf{Indoor} & \textbf{NLOS} & \multicolumn{3}{c}{\textbf{Latent Comp. \& $+3\text{s}$ R@5 (\%)}} & \multicolumn{2}{c}{\textbf{Physical Probes}} \\
    \cmidrule(lr){4-6} \cmidrule(lr){9-11} \cmidrule(lr){12-13}
    & & & \textbf{R@5}$\uparrow$ & \textbf{R@10}$\uparrow$ & \textbf{MedRk}$\downarrow$ & \textbf{Act/Fall}$\uparrow$ & \textbf{Clutter}$\uparrow$ & \textbf{Mid-Occ}$\uparrow$ & \textbf{Drop75}$\uparrow$ & \textbf{Blk$\to$Fut}$\uparrow$ & \textbf{Blk+$3\text{s}$ Tilt}$\downarrow$ & \textbf{UAV Clt Rng/Alt}$\downarrow$ \\
    \midrule
    \multirow{3}{*}{\textbf{Qwen2.5-0.5B} (0.49B)} & \multirow{3}{*}{896 / $r\!=\!24$} & Masked-SFT + LoRA & 0.85 & 1.42 & 356.0 & 91.8/100 & 67.7 & 40.1 & 9.2 & 3.3 & 20.76$^\circ$ (-0.52) & \textcolor{darkred}{\textbf{8.4/3.30}} \\
    & & \cellcolor{litebg}\textbf{+ World-JEPA (Scatter)} & \textcolor{darkred}{\textbf{3.68}} & \textcolor{darkred}{\textbf{6.37}} & \textcolor{darkred}{\textbf{105.0}} & \textcolor{darkred}{\textbf{96.8/100}} & 88.2 & 51.0 & 18.0 & 8.1 & 28.91$^\circ$ (-0.28) & 8.6/3.37 \\
    & & \cellcolor{litebg}\textbf{+ World-JEPA (Span-Euc)} & 3.26 & 6.09 & 147.0 & 77.8/40 & \textcolor{darkred}{\textbf{89.7}} & \textcolor{darkred}{\textbf{61.1}} & \textcolor{darkred}{\textbf{25.2}} & \textcolor{darkred}{\textbf{9.9}} & \textcolor{darkred}{\textbf{13.86$^\circ$ (+0.11)}} & 8.9/3.32 \\
    \midrule
    \multirow{3}{*}{\textbf{Llama-3.2-1B} (1.24B)} & \multirow{3}{*}{2048 / $r\!=\!48$} & Masked-SFT + LoRA & 0.14 & 1.27 & 359.0 & \textcolor{darkred}{\textbf{97.5/100}} & 75.0 & 83.3 & 19.8 & 6.8 & 9.88$^\circ$ (+0.61) & 7.8/1.81 \\
    & & \cellcolor{litebg}\textbf{+ World-JEPA (Scatter)} & 3.68 & 8.36 & \textcolor{darkred}{\textbf{95.0}} & 95.0/80 & 83.8 & 95.8 & 51.4 & 13.7 & 12.84$^\circ$ (+0.77) & 7.8/2.94 \\
    & & \cellcolor{litebg}\textbf{+ World-JEPA (Span-Euc)} & \textcolor{darkred}{\textbf{4.25}} & \textcolor{darkred}{\textbf{8.64}} & 111.0 & 90.4/60 & \textcolor{darkred}{\textbf{88.2}} & \textcolor{darkred}{\textbf{98.4}} & \textcolor{darkred}{\textbf{63.5}} & \textcolor{darkred}{\textbf{18.3}} & \textcolor{darkred}{\textbf{6.31$^\circ$ (+0.89)}} & \textcolor{darkred}{\textbf{7.7/2.42}} \\
    \midrule
    \multirow{3}{*}{\textbf{DeepSeek-R1-1.5B} (1.78B)} & \multirow{3}{*}{1536 / $r\!=\!36$} & Masked-SFT + LoRA & 0.71 & 1.42 & 366.0 & \textcolor{darkred}{\textbf{96.0/100}} & 75.0 & 50.1 & 13.3 & 11.8 & 12.22$^\circ$ (+0.83) & 8.5/3.20 \\
    & & \cellcolor{litebg}\textbf{+ World-JEPA (Scatter)} & 2.69 & 5.67 & 154.0 & 70.8/40 & \textcolor{darkred}{\textbf{85.3}} & 58.8 & 32.4 & 12.3 & 17.96$^\circ$ (+0.07) & 8.3/2.91 \\
    & & \cellcolor{litebg}\textbf{+ World-JEPA (Span-Euc)} & \textcolor{darkred}{\textbf{3.82}} & \textcolor{darkred}{\textbf{8.22}} & \textcolor{darkred}{\textbf{103.0}} & 73.7/40 & 79.4 & \textcolor{darkred}{\textbf{90.1}} & \textcolor{darkred}{\textbf{62.0}} & \textcolor{darkred}{\textbf{16.3}} & \textcolor{darkred}{\textbf{7.01$^\circ$ (+0.83)}} & \textcolor{darkred}{\textbf{8.1/3.75}} \\
    \midrule
    \multirow{3}{*}{\textbf{SmolLM2-1.7B} (1.71B)} & \multirow{3}{*}{2048 / $r\!=\!48$} & Masked-SFT + LoRA & 0.85 & 1.56 & 354.0 & 74.0/40 & 64.7 & 6.4 & 2.0 & 2.5 & 16.23$^\circ$ (+0.42) & \textcolor{darkred}{\textbf{6.4/2.32}} \\
    & & \cellcolor{litebg}\textbf{+ World-JEPA (Scatter)} & \textcolor{darkred}{\textbf{4.25}} & \textcolor{darkred}{\textbf{7.65}} & \textcolor{darkred}{\textbf{110.0}} & \textcolor{darkred}{\textbf{95.7/100}} & 76.5 & 31.7 & 7.1 & 6.9 & \textcolor{darkred}{\textbf{5.61$^\circ$ (+0.92)}} & 8.5/3.44 \\
    & & \cellcolor{litebg}\textbf{+ World-JEPA (Span-Euc)} & 2.69 & 5.24 & 132.0 & 73.8/40 & \textcolor{darkred}{\textbf{80.9}} & \textcolor{darkred}{\textbf{39.7}} & \textcolor{darkred}{\textbf{13.2}} & \textcolor{darkred}{\textbf{10.3}} & 19.64$^\circ$ (0.00) & 8.4/2.67 \\
    \midrule
    \multirow{3}{*}{\textbf{Qwen3-1.7B} (1.72B)} & \multirow{3}{*}{2048 / $r\!=\!48$} & Masked-SFT + LoRA & 0.71 & 1.27 & 336.0 & 88.8/60 & \textcolor{darkred}{\textbf{88.2}} & 47.4 & 11.3 & 7.2 & 11.37$^\circ$ (+0.69) & \textcolor{darkred}{\textbf{8.4/2.57}} \\
    & & \cellcolor{litebg}\textbf{+ World-JEPA (Scatter)} & \textcolor{darkred}{\textbf{4.39}} & \textcolor{darkred}{\textbf{7.79}} & \textcolor{darkred}{\textbf{88.0}} & \textcolor{darkred}{\textbf{95.6/100}} & \textcolor{darkred}{\textbf{88.2}} & 43.2 & 17.0 & 10.2 & \textcolor{darkred}{\textbf{6.32$^\circ$ (+0.80)}} & 8.7/1.20 \\
    & & \cellcolor{litebg}\textbf{+ World-JEPA (Span-Euc)} & 3.68 & 6.80 & 105.0 & 75.6/40 & 60.3 & \textcolor{darkred}{\textbf{72.2}} & \textcolor{darkred}{\textbf{36.5}} & \textcolor{darkred}{\textbf{14.6}} & 8.89$^\circ$ (+0.59) & 8.9/3.40 \\
    \midrule
    \rowcolor{bestbg}\textbf{Qwen3-1.7B + PhysRF} & 2048 / $r\!=\!48$ & \textbf{PhysRF-WorldJEPA (Full Ours)} & \best{9.77} & \best{15.01} & \best{60.0} & \textbf{94.8/100} & \textbf{85.3} & \textbf{61.8} & \textbf{29.5} & \textbf{12.5} & \best{3.37$^\circ$ (+0.96)} & \textbf{7.5/1.89} \\
    \bottomrule
  \end{tabular}}
  \vspace{-0.45em}
\end{table*}

\begin{figure*}[t]
  \centering
  \includegraphics[width=\textwidth]{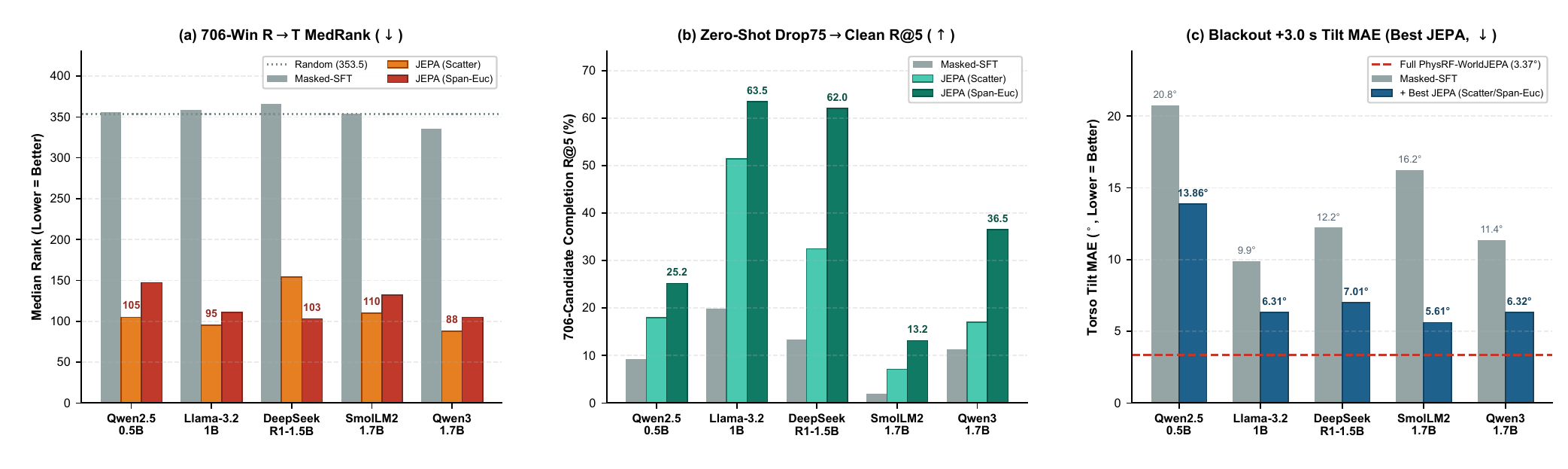}
  \vspace{-0.4em}
  \caption{\textbf{Cross-architecture generalization across five heterogeneous local LLM families (15 zero-overhead models + \texttt{PhysRF} bridge).} Using the identical minimal \texttt{UnifiedRadarEncoder} stem and $\sim\!3\%$ LoRA ($0$ extra inference parameters): \textbf{(a)} all $10/10$ World-JEPA models compress 706-window $\text{R}\rightarrow\text{T}$ median rank from near-random ($336\text{--}366$) into $88\text{--}154$ (best per family $88\text{--}110$); \textbf{(b)} zero-shot extreme $75\%$-drop completion R@5 climbs monotonically (\texttt{SFT} $<$ \texttt{JEPA-Scatter} $<$ \texttt{JEPA-Span-Euc}) across all 5 families, reaching $63.5\%$ on \texttt{Llama-3.2} and $62.0\%$ on \texttt{DeepSeek-R1}; and \textbf{(c)} on blackout $+3.0\,\text{s}$ torso tilt, fixed \texttt{JEPA-Span-Euc} beats SFT on $4/5$ families, fixed \texttt{JEPA-Scatter} on $2/5$, and the better variant per family (plotted) beats SFT across all $5/5$ families ($5.61^\circ\text{--}13.86^\circ$), while adding \texttt{PhysRF} pushes error down to $\mathbf{3.37^\circ}$ ($\mathbf{2.79^\circ}$ Lite).}
  \label{fig:crossllm}
  \vspace{-0.4em}
\end{figure*}

\section{Main Experimental Results}
\label{sec:results}

\subsection{Unified Perception and Global Cross-Modal Retrieval}
\label{sec:res_main1}

Table~\ref{tab:main1} and Fig.~\ref{fig:matrix}a compare PhysRF-WorldJEPA against all 10 external baselines. First, \textbf{non-contrastive SFT and modern PEFT methods are blind to global cross-modal retrieval}: across 1.72B Full FT, Masked LoRA, GaLore, LLaVA+NEFTune, DoRA, PiSSA, and SimPO, $\text{Radar}\rightarrow\text{Text}$ median rank stagnates at $336.0\text{--}367.0$ ($\text{R@10} \le 1.42\%$). Second, \textbf{PhysRF-WorldJEPA achieves the strongest retrieval without sacrificing degraded perception}: Stage~2B reaches a locked-test median rank of $\mathbf{60.0}$ ($\text{R@5}=9.77\%$, $\text{R@10}=15.01\%$)---nearly doubling coupled SigLIP+SFT ($83.0$, $7.65\%$) and improving $\text{R@10}$ by $>10\times$ over DoRA, PiSSA, and LLaVA. Third, \textbf{world modeling eliminates tail collapse under blackouts and clutter} (Fig.~\ref{fig:degradation}): whereas GaLore plunges to $60.8\%$ on \texttt{Drop75} ($20\%$ fall recall) and $32.4\%$ on NLOS blackout, and SimPO drops to $52.9\%$ on cluttered NLOS, PhysRF-WorldJEPA sustains $94.0\%\text{--}96.6\%$ indoor recall, $\mathbf{100\%}$ fall recall across all $8/8$ modes, and $86.8\% / 85.3\%$ NLOS accuracy.

\begin{figure*}[t]
  \centering
  \includegraphics[width=\textwidth]{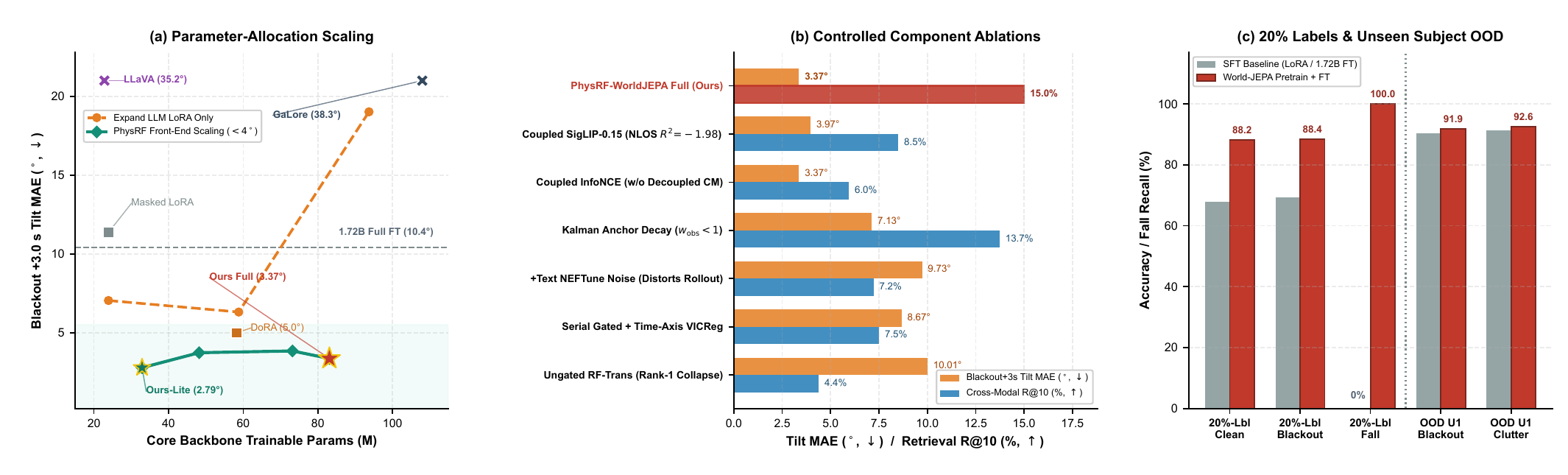}
  \vspace{-0.4em}
  \caption{\textbf{Parameter-allocation scaling law, controlled architectural ablations, and deployment efficiency.} \textbf{(a)} Expanding LLM LoRA rank alone triggers an overfitting U-turn at $5\%$ ($19.02^\circ$ blackout error), whereas allocating capacity to the PhysRF front-end maintains $<4^\circ$ error across scales. \textbf{(b)} Single-factor ablations isolating ungated rank-1 collapse, temporal VICReg pose distortion, NEFTune rollout degradation, Kalman anchor attenuation ($w_{\text{obs}}<1$), and coupled vs.\ decoupled alignment. \textbf{(c)} $20\%$ low-label sample efficiency ($28$ sequences) and zero-shot generalization to unseen subject \texttt{U1} ($608$ windows).}
  \label{fig:ablations}
  \vspace{-0.4em}
\end{figure*}

\subsection{Continuous 3D/NLOS/Pose Rollout and RIS Invariance}
\label{sec:res_probes}

Table~\ref{tab:probes} evaluates zero-shot continuous state readout from the identical LLM terminal causal state $\mathbf{h}_t$ across all methods. Seven of eight external baselines suffer negative $R^2$ on at least one continuous physical domain: LLaVA+NEFTune and GaLore drift to $35.15^\circ$ and $38.30^\circ$ ($R^2<0$) on blackout $+3.0\,\text{s}$ torso tilt (Fig.~\ref{fig:teaser}b); PiSSA+LoRA+ collapses on behind-wall NLOS range ($4.54\,\text{m}$, $R^2=-1.534$); and coupled SigLIP+SFT suffers negative $R^2$ on both 3D UAV altitude ($-1.145$) and NLOS range ($-0.295$). The sole external baseline with positive tri-domain $R^2$, SimPO ($+0.963, +0.090, +0.466$), drops to near-zero UAV altitude $R^2$ ($2.52\,\text{m}$), $52.9\%$ cluttered NLOS classification, and near-random retrieval ($\text{MedRank}=343.0$). In contrast, \textbf{PhysRF-WorldJEPA} is the only model combining strong global retrieval ($\text{MedRank}=60.0$, $\text{R@10}=15.01\%$) with substantial positive $R^2$ across all three domains from $\mathbf{h}_t$: $\mathbf{3.37^\circ}$ ($R^2=+0.961$) on blackout $+3.0\,\text{s}$ tilt ($\mathbf{2.79^\circ}$ on Lite), $\mathbf{1.89\,\text{m}}$ ($R^2=+0.679$; further tightened to $\mathbf{1.19\,\text{m}}$, $R^2=+0.863$ when probing Stage~1 $\bar{\mathbf{x}}_{\text{rf}}$ directly, Fig.~\ref{fig:teaser}f) on cluttered 3D UAV altitude, and $\mathbf{2.04\,\text{m}}$ ($R^2=+0.461$) on behind-wall NLOS $+3.0\,\text{s}$ range, while achieving $\mathbf{0.9976}$ cosine similarity ($\mathbf{100.0\%}$ R@10, $\text{MedRank}=\mathbf{1.0}$) across paired raw-no-RIS vs.\ $+37.4\,\text{dB}$ virtual-RIS NLOS trajectories without pairing supervision.

\subsection{Generalization Across Five Heterogeneous Local LLM Families}
\label{sec:res_crossllm}

A critical question for any LLM-based world model is whether physical grounding depends on a specific LLM family or derives purely from custom front-end layers. Table~\ref{tab:cross_llm} and Fig.~\ref{fig:crossllm} report our strictly controlled \textbf{15-model zero-overhead study} across five diverse $0.49\text{B}\text{--}1.78\text{B}$ families (\texttt{Qwen2.5-0.5B} $D=896$ GQA, \texttt{Llama-3.2-1B} 128K GQA, \texttt{DeepSeek-R1-Distill-1.5B} CoT RL, \texttt{SmolLM2-1.7B} 49K MHA, and \texttt{Qwen3-1.7B}), retaining the minimal \texttt{UnifiedRadarEncoder} stem ($0$ temporal conv/attention layers, $0\,\text{ms}$ inference overhead):
\begin{enumerate}[leftmargin=1.4em,itemsep=1.8pt,topsep=2pt]
  \item \textbf{Consistent 10/10 Retrieval, Completion, \& Rollout Gains:} Standard \texttt{Masked-SFT + LoRA} fails at global retrieval on \emph{every} LLM family ($\text{MedRank} \in [336, 366]$). Without tuning per-family hyperparameters, all 10 zero-overhead World-JEPA runs reduce median rank to $\mathbf{88.0\text{--}154.0}$ (per-family best $88.0\text{--}110.0$; $\text{R@5}$ up to $30.4\times$ on \texttt{Llama-3.2-1B}), while \texttt{Drop75$\to$Clean} completion rises monotonically (\texttt{SFT} $<$ \texttt{JEPA-Scatter} $<$ \texttt{JEPA-Span-Euc}) across all 5 families to $\mathbf{63.5\%}$ on \texttt{Llama-3.2-1B} ($\mathbf{98.4\%}$ \texttt{Mid-Occ}) and $\mathbf{62.0\%}$ on \texttt{DeepSeek-R1-1.5B} ($\mathbf{90.1\%}$ \texttt{Mid-Occ}), alongside $10/10$ gains in $1.5\,\text{s}$-blackout $\to +3.0\,\text{s}$ future latent forecasting, while improving cluttered NLOS accuracy on $4/5$ families ($+10.3\%\text{ to }+22.0\%$) and matching SFT ($88.2\%$) with the best variant on \texttt{Qwen3-1.7B}.
  \item \textbf{Blackout $+3.0\,\text{s}$ Physical Grounding Across Families:} On blackout $+3.0\,\text{s}$ torso tilt (Fig.~\ref{fig:crossllm}c), fixed \texttt{JEPA-Span-Euc} beats SFT on \textbf{4/5 families} ($20.76^\circ \to \mathbf{13.86^\circ}$ on \texttt{Qwen2.5-0.5B}; $9.88^\circ \to \mathbf{6.31^\circ}$ on \texttt{Llama-3.2-1B}; $12.22^\circ \to \mathbf{7.01^\circ}$ on \texttt{DeepSeek-R1-1.5B}; $11.37^\circ \to 8.89^\circ$ on \texttt{Qwen3-1.7B}), fixed \texttt{JEPA-Scatter} beats SFT on \textbf{2/5 families} ($16.23^\circ \to \mathbf{5.61^\circ}$ with $40\%\to\mathbf{100\%}$ fall recall on MHA \texttt{SmolLM2-1.7B}; $11.37^\circ \to \mathbf{6.32^\circ}$ and $1.20\,\text{m}$ UAV clutter altitude on \texttt{Qwen3-1.7B}), and choosing the better World-JEPA recipe per family outperforms SFT on all \textbf{5/5 families}.
  \item \textbf{Causal Observability vs.\ Euclidean Calibration Synergy:} Contrasting the two zero-overhead variants reveals why both mechanisms matter: \texttt{JEPA-Span-Euc} (Eq.~\eqref{eq:spaneuc}) calibrates high-norm \texttt{[MASK]} sinks in instruction and CoT-distilled LLMs ($12.84^\circ\to 6.31^\circ$ on \texttt{Llama-3.2}; $17.96^\circ\to 7.01^\circ$ on \texttt{DeepSeek-R1}), whereas \texttt{JEPA-Scatter} preserves sparse frame cues during trailing $1.5\,\text{s}$ human falls for supervised $\mathcal{L}_{\text{CE}}$ ($95.0\%\text{--}96.8\%$ action recall). Pairing this universal training foundation with our \texttt{PhysRF} front-end and Stage~2B head (bottom row of Table~\ref{tab:cross_llm}) further advances retrieval ($\text{MedRank}=60.0$) and blackout forecasting ($\mathbf{3.37^\circ}$).
\end{enumerate}

\begin{table}[t]
  \centering
  \caption{\textbf{Controlled architectural \& objective ablations on \texttt{Qwen3-1.7B}.} Each row modifies one design axis relative to full PhysRF-WorldJEPA.}
  \label{tab:ablation}
  \scriptsize
  \setlength{\tabcolsep}{3.2pt}
  \renewcommand{\arraystretch}{1.10}
  \resizebox{\columnwidth}{!}{\begin{tabular}{l c c c c}
    \toprule
    \textbf{Ablation Variant} & \textbf{MedRk}$\downarrow$ & \textbf{R@10}$\uparrow$ & \textbf{Blk+$3\text{s}$}$^\circ\downarrow$ & \textbf{NLOS $R^2$}$\uparrow$ \\
    \midrule
    1) Ungated Serial RF Transformer & 208.0 & 4.39 & 10.01 & $-0.169$ \\
    2) Serial Gate + Time-Axis VICReg & 101.0 & 7.51 & 8.67 & $-0.860$ \\
    3) Expand LLM 5\% Only (w/o PhysRF) & 88.0 & 7.22 & 19.02 & $-0.571$ \\
    4) +Text NEFTune ($\alpha=3.0$) & 82.0 & 7.22 & 9.73 & $-0.412$ \\
    5) Kalman Anchor Gate ($w_{\text{obs}} < 1$) & \best{55.0} & 13.74 & 7.13 & $-0.605$ \\
    \midrule
    6) Coupled SigLIP ($w=0.25$) & 83.0 & 7.51 & 4.50 & $-1.324$ \\
    7) Coupled SigLIP ($w=0.15$) & 81.0 & 8.50 & 3.97 & $-1.982$ \\
    8) Coupled InfoNCE (w/o Stage 2B) & 107.0 & 5.95 & \best{3.37} & \best{+0.461} \\
    \midrule
    \rowcolor{bestbg}
    \textbf{9) PhysRF-WorldJEPA (Full Ours)} & 60.0 & \best{15.01} & \best{3.37} & \best{+0.461} \\
    \bottomrule
  \end{tabular}}
  \vspace{-0.6em}
\end{table}

\section{Diagnostic Ablations and Deployment Value}
\label{sec:ablations}

Table~\ref{tab:ablation} and Fig.~\ref{fig:ablations} isolate each architectural axis:
(1) \textbf{Parameter-Allocation Law:} Expanding LLM LoRA to $5\%$ without \texttt{PhysRF} (Row~3) causes blackout tilt error to U-turn to $19.02^\circ$, whereas keeping LoRA at $1\%\text{--}3\%$ and allocating $9.2\text{M}\text{--}24.5\text{M}$ to \texttt{PhysRF} cuts error $5.6\times$ to $3.37^\circ$ ($2.79^\circ$ Lite).
(2) \textbf{Collapse Prevention \& $1.0\times\mathbf{x}_0$ Clock Anchor:} Ungated serial RF attention collapses (Row~1, $23.5\%$ NLOS); temporal-axis VICReg distorts fallen poses (Row~2, $8.67^\circ$); NEFTune perturbs rollout (Row~4, $9.73^\circ$); and convex Kalman gating ($w_{\text{obs}}<1$, Row~5) attenuates blackout clock indices $\mathbf{p}_{16:30}$ ($3.37^\circ \to 7.13^\circ$).
(3) \textbf{Seesaw Diagnosis \& Decoupled Alignment:} Coupled SigLIP at both $w=0.25$ and $w=0.15$ (Rows~6--7) drives NLOS $R^2$ deeply negative ($-1.324, -1.982$) and worsens blackout tilt ($4.50^\circ, 3.97^\circ$); removing SigLIP coupling (\texttt{InfoNCE}, Row~8) restores physical readout ($3.37^\circ$, NLOS $R^2=+0.461$) at the cost of retrieval ($\text{R@10}=5.95\%$, $\text{MedRank}=107.0$), whereas freezing the backbone and training only the Stage~2B decoupled head (Row~9) lifts $\text{R@10}$ to $\mathbf{15.01\%}$ ($\text{MedRank}=\mathbf{60.0}$) with zero physical degradation.
(4) \textbf{Low-Label \& Unseen-Subject Generalization:} Under just \textbf{20\% labeled sequences} (Fig.~\ref{fig:ablations}c), Stage~1 EMA-JEPA pretraining boosts blackout macro-recall by $\mathbf{+36.6\%}$ ($54.5\%\to 91.1\%$; micro window accuracy $69.3\%\to 88.4\%$) and rescues fall recall from $\mathbf{0\% \to 100\%}$; on 608 separate held-out windows of unseen subject \texttt{U1}, PhysRF-WorldJEPA achieves $\mathbf{92.9\% / 91.9\% / 92.6\%}$ (clean/blackout/clutter), surpassing 1.72B Full FT ($91.5\% / 90.5\% / 91.5\%$).

\section{Limitations and Conclusion}
\label{sec:conclusion}

\paragraph{Limitations.}
Because stationary postures (\texttt{standing}, \texttt{sitting}, \texttt{hovering}) are symmetric under time reversal, we omit unconditional time-reversal margin penalties (which degrade clean accuracy from $96.7\%$ to $88.3\%$); velocity-gated temporal arrow losses on high-motion segments, scaling to larger language backbones, and closed-loop active RIS beamforming control remain promising directions for future work.

\paragraph{Conclusion.}
We presented \textbf{PhysRF-WorldJEPA}, uniting an unattenuated $1.0\times$ clock anchor, parallel dual-scale RF kinematics with sample-axis Batch-VICReg, two-step $\text{Denoise}\rightarrow\text{Rollout}$ EMA-JEPA with scale-normalized Euclidean anchoring, and stop-gradient manifold-decoupled alignment. Validated against 10 modern baselines and across 15 models spanning five heterogeneous local LLM families, our framework resolves the Geometry--Semantics Seesaw and establishes a backbone-agnostic paradigm for electromagnetic world modeling with LLMs.

\bibliographystyle{unsrtnat}
\bibliography{references}

\end{document}